\documentclass{article}

\usepackage{microtype}
\usepackage{graphicx}
\usepackage{subfigure}
\usepackage{booktabs} 

\usepackage{hyperref}

\usepackage[accepted]{icml2023}

\usepackage{amsmath}
\usepackage{amssymb}
\usepackage{mathtools}
\usepackage{amsthm}

\usepackage[capitalize,noabbrev]{cleveref}

\theoremstyle{plain}

\theoremstyle{definition}

\theoremstyle{remark}

\icmltitlerunning{Leveraging Contextual Counterfactuals Toward Belief Calibration}

\begin{document}

\twocolumn[
\icmltitle{Leveraging Contextual Counterfactuals Toward Belief Calibration}



\icmlsetsymbol{equal}{*}

\begin{icmlauthorlist}
\icmlauthor{Qiuyi (Richard) Zhang}{yyy}
\icmlauthor{Michael S. Lee}{comp}
\icmlauthor{Sherol Chen}{yyy}
\end{icmlauthorlist}

\icmlaffiliation{yyy}{Google Deepmind}
\icmlaffiliation{comp}{The Robotics Institute, Carnegie Mellon University, Pittsburgh, PA, USA}

\icmlcorrespondingauthor{Michael S. Lee}{ml5@andrew.cmu.edu}

\icmlkeywords{Machine Learning, ICML}

\vskip 0.3in
]



\printAffiliationsAndNotice{}  

\begin{abstract}
Beliefs and values are increasingly being incorporated into our AI systems through alignment processes, such as carefully curating data collection principles or regularizing the loss function used for training. However, the meta-alignment problem is that these human beliefs are diverse and not aligned across populations; furthermore, the implicit strength of each belief may not be well calibrated even among humans, especially when trying to generalize across contexts. Specifically, in high regret situations, we observe that contextual counterfactuals and recourse costs are particularly important in updating a decision maker's beliefs and the strengths to which such beliefs are held. Therefore, we argue that including counterfactuals is key to an accurate calibration of beliefs during alignment. To do this, we first segment belief diversity into two categories: subjectivity (across individuals within a population) and epistemic uncertainty (within an individual across different contexts). By leveraging our notion of epistemic uncertainty, we introduce `the belief calibration cycle' framework to more holistically calibrate this diversity of beliefs with context-driven counterfactual reasoning by using a multi-objective optimization. We empirically apply our framework for finding a Pareto frontier of clustered optimal belief strengths that generalize across different contexts, demonstrating its efficacy on a toy dataset for credit decisions.
\end{abstract}

\section{Introduction}

With the rapid development of AI and ML and its pervasive use throughout all segments of society, there has been an increasing awareness of the necessity to assess and remediate the societal effects of these emerging technologies. Because AI systems and their design will always encode some values or beliefs about the world, one major challenge to AI alignment is that some commonly accepted values have been shown to be fundamentally incompatible \citep{friedler2021possibility}. 

For example, underlying value differences has led to an overwhelming variety of mathematical formulations of fairness metrics. But fairness impossibility theorems demonstrate that it is in fact mathematically impossible to simultaneously satisfy even the three common and intuitive definitions of fairness - demographic parity, equalized odds, and predictive rate parity \citep{miconi2017impossibility}. Furthermore, individual fairness dictates that individuals who are similar should be treated similarly with respect to a specific outcome \citep{dwork2012fairness}. Yet group fairness states that demographic groups should, on the whole, receive similar decisions \citep{binns2020apparent}. And even if group fairness were the agreed upon standard, research has shown that even between groups, there are disparate notions of fairness and toxicity which are not universally generalizable. For example in the context of forum content moderation, each community implicitly subscribes to a different social contract, so toxicity for one community is not the same for another \citep{goyal2022your}. Although belief and value disagreement has increasingly become a critical issue in social alignment of new AI technology, it seems difficult to model, let alone resolve, these conflicts.

One way forward is spurred on by recent progress in large language models (LLMs), which has opened the door for AI researchers to develop general values via a set of language-based contracts (e.g. constitutions or rule sets) that are simultaneously-held without the need for direct (and perhaps conflicting) mathematical translation. Specifically, the technique aims to guide the training of an AI assistant through a list of rules and principles that guide the labeling of training data \cite{bai2022constitutional, glaese2022improving, thoppilan2022lamda}. 
Though rule-assisted methods have shown increased ability to provide harmless and helpful feedback, they each differ in their respective rule-sets due to subjective notions of good. Moreover, even if some set of constitutional tenets are generally accepted, each specific belief is held to varying degrees of strength among groups of individuals. And even in an individual, it is well known that belief uncertainty exists and is often not well calibrated, leading to the popular overconfidence bias \cite{moore2008trouble}. This creates what we call the {\it meta-alignment problem}: { Even if a general set of AI alignment values are identified, how should the model calibrate the strength of each value for beneficial societal impact?} 

Beneficial social impact is an inherent multi-faceted objective and has been captured by various dimensions of good, such as fairness, privacy, explanability, transparency, etc. Fairness efforts, such as those that equalize false positive rates for recidivism prediction \cite{dieterich2016compas}, attempt to reduce bias and toxicity in algorithmic decisions or model outputs \cite{dwork2012fairness}. In a related vein, privacy efforts attempt to protect the sensitive attributes of an individual \cite{dwork2008differential} and other recent efforts like model cards \citep{mitchell2019model} try to provide transparency into the model mechanisms as well as its benchmarked performance. 

In our case, we focus on an overlooked yet crucial dimension of social benefit: cost-effective positive outcome generation, which is captured by the study of counterfactuals and recourses. Going back to the example of forum moderation, note that while high toxicity posts should be banned in an effort to maintain social order, it also can impede on the freedom of speech or expression. In light of this, it is thereby critical for the moderation system to minimize -- or at least calibrate -- the remediation effort it requires for an individual to post their honest opinions. Indeed, on another extreme, an AI moderation system that makes it impossible or extremely costly for a specific individual to express their opinions could be seen as a form of censorship and a violation of free speech \cite{massaro2015siri}.

To address this, recent work has utilized counterfactual reasoning to consider not only different outcomes in response to different hypothetical input features \cite{wexler2019if} or model choice \cite{bui2022counterfactual} to minimize bias or overfitting \cite{gomez2021advice}, but also in response to changes in features grounded in an individual in the form of recourses \cite{gomez2020vice}. For example, linear models for credit determination has been analyzed for in the light of actionable recourse \citep{ustun2019actionable}, showing that given a negative outcome, the ability for an individual to induce counterfactual results often varies drastically based on model design. Therefore, recourse should be included as a critical factor to inform alignment efforts on producing positive societal impact.


Furthermore, counterfactuals have also long been studied in the social sciences as a tool in human decision making, for both learning from past outcomes \cite{roese1997counterfactual}, and weighing future outcomes to minimize future regret \cite{simonson1992influence}. As we continue to build systems that ultimately strive to emulate human decision making, we argue for leveraging counterfactual reasoning to calibrate to human beliefs. And as we deploy these systems in high impact domains such as finance \cite{murawski} and criminal justice \cite{hao}, we argue for leveraging counterfactuals to adjust to foreseen and observed consequences in a principled manner both prior to and throughout deployment. 

\subsection{Our Contributions}

In this paper, we provide a framework for tackling the meta-alignment problem of calibrating diverse human beliefs and values for maximizing beneficial societal impact, as defined from a counterfactual perspective. Building off of previous works that focus on outcome diversity from subjectivity among human populations, we argue that counterfactuals plays a critical role in calibrating and inducing belief diversity. Specifically, studies have shown the immense power that counterfactual reasoning (and anticipated regret) has to shape one's beliefs \textit{prior} to decision making, such as for avoiding risky behaviors like reckless driving. Furthermore, we note that the context in which these counterfactuals are analyzed is also critical for appropriately grounding the strength to which beliefs are upheld in decision making. For example, given the same classification task, human raters are shown to give more lenient labels when they are told that those labels are used for more severe penal punishments, suggesting that humans implicitly include the recourse cost in their calibration process.

In modeling belief diversity, we partition belief diversity into two categories: subjectivity and epistemic uncertainty. Subjectivity acknowledges that different individuals within a population maintain different sets of salient values to varying degrees. And epistemic uncertainty acknowledges that even an individual upholds these values to varying degrees across different contexts. The latter humbly acknowledges that an individual's beliefs are generated by only a myopic view of the world, a view that is changing and shifting as that person constantly gains new insights and finds themselves in different circumstances \cite{schwartz2006basic}. Specifically for epistemic uncertainty, we emphasize that counterfactuals help illuminate an important context in inducing belief diversity, as alignment values can change based on the task-dependent consequences.

\begin{figure}[tbh]
\centering
\includegraphics[width=\columnwidth]{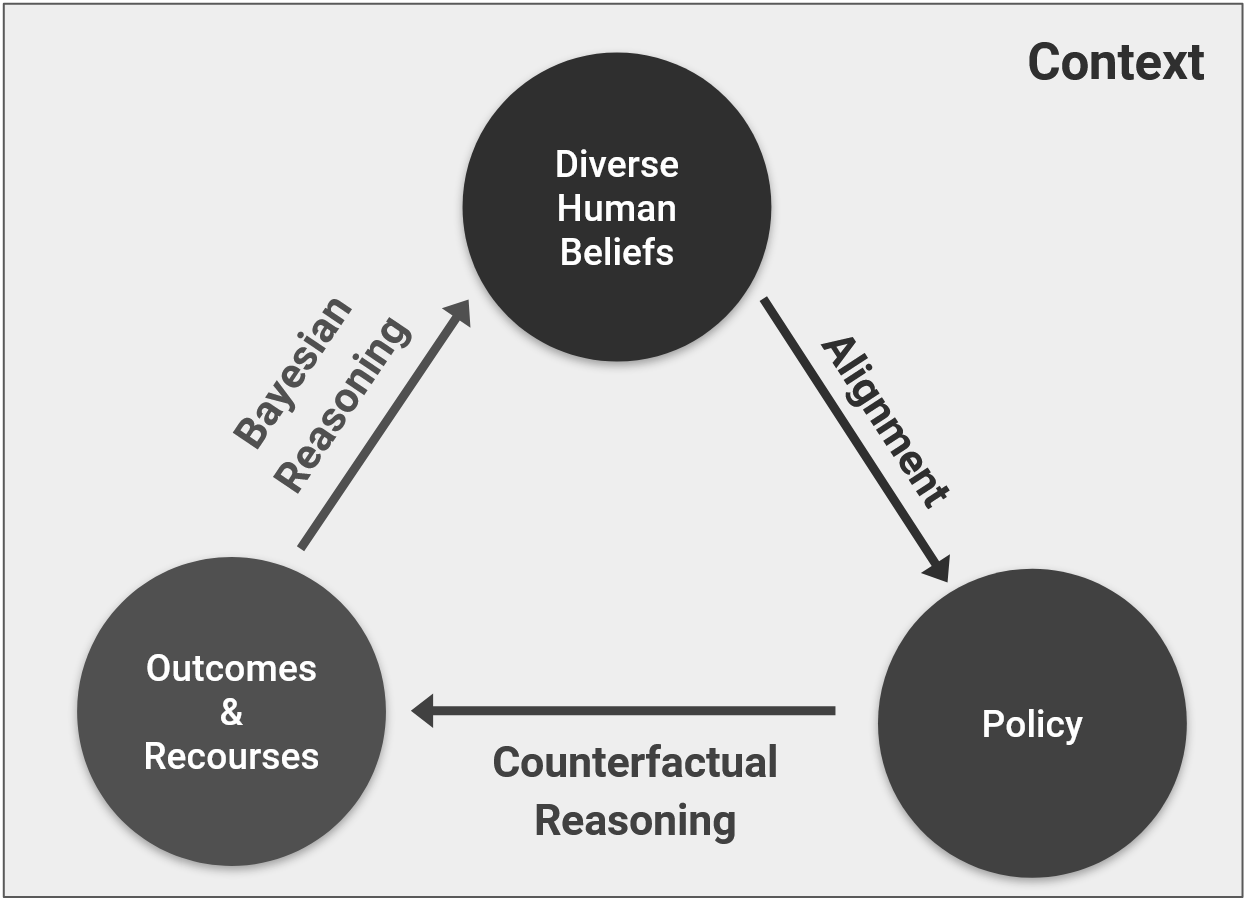} 
\caption{Belief calibration cycle. We propose to leverage counterfactual reasoning over the outcomes and recourses to calibrate towards beliefs that ensure outcomes and recourses that are beneficial toward decision subjects. Finally, we emphasize that this calibration occurs within a specific context that influences the beliefs, policy, outcomes and recourses and vice versa. }
\label{fig:calibration_loop}
\end{figure}

Finally, in order to solve the two-fold problem of marrying alignment and calibration, we introduce a framework for understanding the belief update and calibration feedback loop, which allows for population and individual-level beliefs to not only govern decision-making policies but also to be updated based on the observed outcomes and recourses. This closed-loop cycle that leverages counterfactual reasoning to continuously calibrated the beliefs incorporated in our decision making algorithms, ultimately leading to desired outcomes and recourses (see  Fig. \ref{fig:calibration_loop}). We leverage our Bayesian framework to not only induce outcome diversity but also a distributional understanding of counterfactual recourses. We apply counterfactuals in multiple contexts to solve a multi-objective problem that attempts to model the multiple facets of societal impact. Unlike previous works in Bayesian modeling and alignment, we acknowledge that both the probabilistic accuracy of the predictions and the societal consequences of such predictions are critical for joint calibration.

\begin{figure}[tbh]
\centering
\includegraphics[width=\columnwidth]{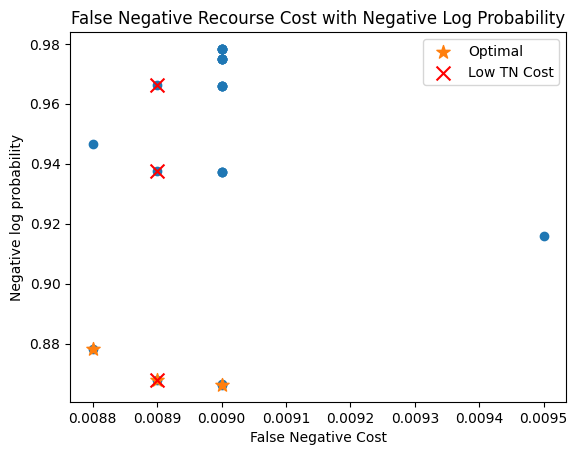}
\resizebox{0.8\columnwidth}{!}{%
\begin{tabular}{ccccc}
\toprule 
$\sigma$ &  $\lambda$ & FN Cost & TN Cost &  Log-Prob \\
\midrule
10 & 0.01 & 0.0088 & 0.0189 & 0.878 \\
10 & 10 & 0.009 & 0.0189 & 0.866 \\
\bottomrule
\end{tabular}
}
\caption{Belief calibration on the \texttt{credit} dataset by minimizing average false negative recourse cost and negative log probability, via a scatter plot along with a table of Pareto optimal points. To disincentivize risky behavior, all beliefs corresponding to a higher risk of credit misuse (i.e. low true negative recourse cost) are removed, leaving only two optimal beliefs.}
\label{fig:tn}
\vspace{-4mm}
\end{figure}

We apply our contextual calibration approach on the \texttt{credit} dataset and introduce novel alignment contexts and measures of counterfactual recourse cost, such as the true negative and false negative recourse cost. For example, a loan business may wish to calibrate beliefs on the noise ($\sigma$) and sparsity ($\lambda$) model parameters by simultaneously 1) maximizing accuracy, 2) decreasing the recourse cost of loans that were falsely denied (false negative), while at the same time 3) increasing or thresholding on the recourse cost of loans that are justifiably denied (true negative). The last objective, while novel and perhaps counterintuitive in counterfactual cost calibration, is important for sensible alignment since access to free credit is not always desirable and costly consequences are critical for disincentivizing risky behavior. From our calibration experiments in this context, we find that only a few settings on belief priors on noise and sparsity lead to optimal outcomes in all three objectives (see Fig~\ref{fig:tn}).

While there are multiple contexts and objectives, we find that not many of the suggested belief strengths are actually on the Pareto frontier. This suggests that although diversity in counterfactual contexts may lead to an explosion of valid beliefs; in reality, there are only a small cluster of optimal calibrated beliefs. We empirically observe that diverse contextual information does in fact drive diversity in optimal belief strengths, often favoring more mild beliefs. Furthermore, our belief calibration framework reveals surprising trends in the task at hand, such as evidence that lower feature regularization does not always lead to lower recourse cost, or that increased credit leniency can help both predictive and social alignment.

\section{Subjectivity and Related Works}

\subsection{Early Efforts}

Early AI approaches of symbolically modeling beliefs often divided the space into decisions that are either rational or irrational \cite{russel2010}, also referred to as cold and hot cognition \cite{tomkins1963computer}. \citet{abelson1963computer} defined the problem as follows: 

\begin{quote}
Is it possible to specify a realistic model for attitude change and resistance to change in sufficient process detail so that a computer could simulate it? [...] One might even speak of attitudinal problem-solving, wherein the individual is confronted with a challenge to his belief system and the “problem” he must solve is, “What am I to believe now?”
\end{quote}

The major focus here is on cognitive processes, but Abelson proposes to explore some limited relations between cognition and affect. Specifically, within the context of attitudes and attitude changes, one might hope to develop a simulation model which would do for hot cognition what others have done for cold cognition. Implicit in this formulation is the idea that beliefs are fluid and are necessary to be updated to match the changing world.

These AI theorists posed that models of human intelligence were anchored to their belief systems, and that humans employed various techniques (like rationalization) to avoid changing the values of these systems, unlike the objective goals of winning a chess match \cite{abelson1965computer}. Systems like the Goldwater Machine \cite{wardrip2009expressive}, POLITICS \cite{carbonell1978politics}, Terminal Time \cite{mateas2000generation}, and RoleModel \cite{chen2010rolemodel} were developed towards the pursuit of representing subjectivity (often due to affective and ideological differences) within humans.

\subsection{Modern Efforts} 

As foundational as the Turing Test is among computational thought experiments \cite{turing1956can}, believability (or distinguishability) of human-like behavior has been a debatable measure of AI achievement \cite{searle2009chinese}. Recent advancements in LLMs, while trained to satisfy an objective score, can now believably represent diverse systems of beliefs and are a continuation of modeling the aforementioned ``hot" aspects of our cognition, including the (1) subjectivity of each belief-system, and (2) the gravity that our myopia or epistemic miscalibration creates. Much work is focused on downstream remediation of more desireable model outputs, such as reward modeling from human feedback, or Reinforcement Learning from Human Feedback (RLHF) \citep{ouyang2022training}. Such approaches aim to bridge  challenges in aligning to the nuances of human motivation. RLHF, for example, has shown promising results by directly providing feedback from human teachers that permits AI to learn with broader perspective and greater efficiency. However, such methods are nonetheless subject to the inherent biases of human behavior and focus mainly on predictive outcomes or increased personalization \citep{ziegler2019fine}, rather than the consequences of such outcomes.

In light of this, as AI systems are technically more advanced, we posit that the tenuous dividing line between humans and AI, will hinge on analyzing the implicitly held values of an individual and the ethical consequences of such beliefs, as well as their ability to update their beliefs in a rational or irrational way. In some ways, the ability of an AI system to solve the meta-alignment problem of aligning the diversity of human values themselves is a modern day extnesion of Alan's Turing test.

\section{Belief Calibration via Contextual Counterfactuals}

Counterfactual reasoning traditionally considers how altering a factual antecedent to an past event or decision may have lead to a subsequent change in outcome. In synthesizing the early research, \citet{roese1997counterfactual} argues for a functional view of counterfactual thinking as a powerful tool for reasoning over past behavior to improve future outcomes. For example, counterfactuals are more likely to be generated in response to negative situations \cite{sanna1996antecedents}, negative emotions \cite{roese1997counterfactual}, and repeated problems \cite{markman1993mental}). 

\subsection{Counterfactual Reasoning Calibrates Beliefs}

As we design algorithms that increasingly impact peoples' lives, we argue that the community should not only leverage counterfactuals to continuously learn from realized outcomes, but also that that they are a critical component in the re-calibration of our beliefs even prior to deployment. Studies have demonstrated the power that counterfactual reasoning (and anticipated regret) has to shape one's beliefs \textit{prior} to decision making. \citet{parker1996modifying} showed that out of videos that showing different possible outcomes of driving over the speed limit, it was specifically videos that focused on regret that most significantly changed their beliefs about unsafe driving. Similar results of a priori counterfactual reasoning affecting decision making can be found in consumers making safer choices in purchasing \cite{simonson1992influence} and unwillingness to trade identical lottery tickets for additional money, likely influenced by loss aversion toward regret that of trading away a winning ticket \cite{bar1996people}. 

As proposed by Roese, the power of counterfactual reasoning to promote changes in behavior lies especially in learning (post-hoc) or protecting against high consequence decisions (a priori). In 2016, Jeff Bezos popularized two types of decisions -- Type 1 decisions that are consequential and effectively irreversible, in contrast to Type 2 decisions that are reversible -- and argued that ``great deliberation and consultation'' are warranted only for the former \cite{bezos}. We instead build on this framework by recognizing both consequence and reversibility both lie on spectrums and argue for on-going counterfactual reasoning in a calibration loop (see Fig. \ref{fig:calibration_loop}) over expected and outcomes and recourses (or reversibility) as we build algorithms that increasingly impact people's lives. 

\subsection{Contextual Grounding Diversifies Beliefs}

Though counterfactual reasoning is powerful and can help calibrate beliefs by closing the loop, the context in which it is done is critical for how strongly a belief should be upheld. Context has long been acknowledged as being key for appropriately grounding decision making in  affective computing \cite{pantic2006human} and more recently in transparency and explainability \cite{ehsan2021expanding}, and can be broadly captured by the notion of the 5W's that surround a decision, which are `who' the involved parties are, `what' the task is, and `where', `when', and `why' the decision was made. 

We illustrate the importance of grounding counterfactual reasoning in context through a recent study that highlighted the impact that perceived contextual consequences can have on changing human labeling beliefs. They found that humans will significantly differ in their labels of the same data depending on whether it is simply an exercise in factual description (e.g. whether a dog simply looked aggressive) or an exercise in normative judgment (e.g. whether a dog looked aggressive, and therefore violated an apartment's policy) \cite{balagopalan2023judging}. In discussing their results, the authors importantly note that participants may have attached \textit{``a different cost to their judgments in the two conditions: Getting a decision wrong factually is just a matter of describing the world incorrectly. Getting it wrong normatively is a matter of potentially doing harm to another human.''} We posit that the `what' of data labeling was shared between the two conditions with the identical set of features (e.g. aggression), but they differed in the `why' (i.e. for description or judgment of a rule violation), and thus participants likely arrived at different counterfactuals of unequal consequence for the two conditions and adjusted their beliefs and labels accordingly. 

Though different contexts may yield different counterfactuals that will influence human beliefs accordingly, shared contexts can conversely benefit from similar counterfactual reasoning that can yield a shared bias in beliefs. The aforementioned study by \citet{balagopalan2023judging} also notes that participants in the judgment condition (or context) consistently showed ``benefit of the doubt'' in choosing not to flag close calls. Interestingly, this sentiment directly echoes the legal and judicial bias in the United States to usually favor the defendant, captured by the famous adage of "innocent until proven guilty" \cite{baradaran2011restoring}. Presumption of innocence is one of the long standing principles can be traced back to the religious and constitutional beliefs that founded America and aim to protect defendants from negative legal ramifications often with little recourse. Thus we may leverage operating in a shared context to bias beliefs in a shared direction that supports positive outcomes. 

To conclude, we emphasize that people naturally use counterfactuals to assist them in calibrating their beliefs and decisions, especially in high stakes contexts with potentially negative outcomes. As we increasingly automate aspects of decision making that have significant ramification on people's livelihoods, we underscore the importance of also utilizing contextually-grounded counterfactual reasoning to calibrate the beliefs underlying our algorithms to help ensure beneficial outcomes for decision subjects. 


\section{Modeling Diversity of Beliefs}

In this section, we give a Bayesian framework for modeling belief diversity mathematically and applying notions of social calibration for maximum impact. In Bayesian modeling, the idea of a belief is encapsulated by the prior distribution of a set of variables, as a proxy for the inductive and value biases of the model trainer. Generally, we extend this notion by also incorporating the choice of likelihood loss into the belief system structure, thereby reducing the beliefs and values of the trainer to be encapsulated more holistically by the posterior, or the trained model parameters. We note that policy beliefs, which guides the ultimate use of the model to make value-based judgments and decisions, are also integral but will not be mentioned here.

Specifically, let $(x_i, y_i) \in \mathcal{D}$ be datapoints and for model architecture parameterized by trainable parameters $w$, we see that $f_w(x)$ is typically trained by solving $w^\star = \arg \min_{w} \mathcal{L}(f_w(x))$, where $\mathcal{L}$ is some chosen objective with possibly many forms of regularization or penalties. The choice of objective, and also the training method, usually incorporates notions of the trainer's belief to guide the parameters suitably, such as the preference for robustness in adversarial training \cite{gu2014towards}, fairness-aware training via regularization on specific statistical parity based scores \cite{kamishima2011fairness}, or even the bias towards sparse solutions for ease of understanding and visualization \cite{rasmussen2012tutorial}. 

\subsection{Subjectivity vs Epistemic Uncertainty} 

Different belief principles among large groups can be generally encapsulated by different loss functions $\mathcal{L}_1,..., \mathcal{L}_k$, leading to disparate trained parameters $w_1, ... , w_k$. Oftentimes, these loss functions are a combination of fundamental beliefs, with its corresponding loss $\ell_i$, as represented by the prior or the likelihood: $\mathcal{L} = \lambda_1\ell_1 + \lambda_2\ell_2 + ... + \lambda_k \ell_k$, where $\ell_i$ are fundamental beliefs. For example, Bayesian LASSO \cite{park2008bayesian} trades off between the belief that prioritizes accuracy vs the belief for sparsity using some regularization parameter. Therefore, while subjectivity is the study of separate fundamental beliefs, we crucially note the strength to which each belief is expressed, or equivalently its certainty, is tuned as the study of epistemic uncertainty. 

We apply Bayesian modeling to understand how uncertainty in beliefs translate over, via the model and its training process, to uncertainty in trained parameters, and ultimately to outcomes and recourses, where we focus on the former connection in this section. For simplicity, assume that $f_w(x) =  \sum_i w_i \phi_i(x)$, where $\{\phi_i\}_{i=1}^k$ is some $k$ dimensional feature map, with possibly $k = \infty$. Traditional Bayesian models can be viewed as transforming some prior beliefs about $w$ to a posterior distribution over $w$, and is therefore the first step to understand the distribution over outcomes. 

Generally, given some prior distribution $p(w)$ and by incorporating some noise beliefs about our observations, $y = f_w(x)  + \epsilon$, where $\epsilon \sim \mathcal{N}(0, \sigma^2 \mathbb{I})$, we see that the posterior is derived by applying Bayes rule:
\begin{align*}
    p(w | x_i, y_i) &\propto p (y| x_i, w) p(w) \\
    &\propto  \exp\left(-\frac{1}{2\sigma^2}\|y - f_w(x)\|^2\right)p(w) \\
    &\propto \exp\left(-\frac{1}{2\sigma^2}\|y - \Phi(x)w\|^2\right)p(w)
\end{align*}
where $\Phi(x)$ is the matrix of $\phi_i(x)$. Therefore, we see that our likelihood is in fact a Gaussian with an inverse covariance matrix given by what is popularly known as the kernel matrix $K = \Phi(x)^\top\Phi(x)$, where for two input points $x, x' \in \mathbb{R}^d$, $K(x, x') = \sum_i \phi(x)\phi_i(x')$. Now, with the Gaussian prior $p(w)$, the resulting posterior is also Gaussian and is the basis of the popular Gaussian process regression \cite{rasmussen2010gaussian}. 

Even more generally, the noise model and the corresponding loss function may not correspond to a simple Gaussian distribution, as is often the case in classification; and furthermore, the prior may have a complicated expression that incorporates sophisticated beliefs. Therefore, the posterior is often not well analytically expressed, but Bayesian inference methods can extend to these regimes via sampling or approximation methods, such as MCMC and variational inference, usually according to the distribution $p(w | x_i, y_i) \propto \exp(-\mathcal{L}(f_w)) p(w)$ \citep{dellaportas2002bayesian, salimans2015markov}. 


\subsection{Linear Regression}

Since our model choice is targeted for maximal interpretability and explainability, we will focus on simplest notion of linear regression with $f_w(x) = w^\top x$ with $w \in \mathbb{R}^d$, in which case there is a direct correspondence from parameter to feature. Let us consider a per-feature prior on the parameters given by a general multi-variate Gaussian $p(w) = \mathcal{N}(m_p, \Sigma_p)$. Then applying Bayesian inference implies that the posterior is 
$$p(w | x_i, y_i ) \sim \mathcal{N}(  w_{post}  , A^{-1}) $$
where $A = \sigma^{-2} XX^\top + \Sigma_p^{-1}$ and $w_{post} = \sigma^{-2}A^{-1}Xy
$, where $X, y$ are the stacked matrices of $x_i, y_i$. 

Note that even this simple model, there are two classes of beliefs that are captured using the hyperparameters of posterior loss, parameterized by the 1) noise scale $(\sigma)$ and 2) the regularization prior of the parameters $(\Sigma_p)$. The noise scale measures the strength the belief that our data is inherently noisy, with $\sigma \to \infty$ implying that our data is completely random. The regularization prior $\Sigma_p$, when it is a diagonal matrix, represent a prior belief on how much features should be considered when making credit decisions, where $(\Sigma_p)_{ii} \to \infty$ implies that the feature $i$ should not be used at all. 

Typically, these hyperparameters are tuned using some sort of predictive calibration measures, such as the marginal loglikelihood. Automatic relevance determination (ARD) is a standard procedure for tuning Gaussian Process hyperparameters, with $\sigma, \Sigma_p$ as the observation noise and length scale parameters \cite{rasmussen2010gaussian}. However, while such measures give rise to accurate and well-calibrated models, they do not generally consider the downstream societal effects of these choices.

\section{Calibration and Alignment}

The two-fold problem of marrying alignment and calibration starts from the observation that 1) standard Bayesian calibration of hyperparameters is done without alignment to social beliefs and 2) mathematical alignment of models and policies to beliefs is often accomplished without calibration. Essentially, we either only calibrate to probabilistic measures of predictive success or to overgeneralized measures of ideal social outcomes, but not both. In this section, we give a preliminary solution to both issues by using a multi-objective approach for calibration and alignment, from the lens of counterfactual reasoning. Specifically, we consider the \texttt{credit} classification task that was used previously in the study on actionable recourse by linear regression \citep{ustun2019actionable}. We focus on experimental results, where we consider the calibration of a specific individual belief of feature-based weights via both outcome and recourse-based metrics, in multiple contexts.

The \texttt{credit} classification task aims to predict credit default using $d = 17$ features derived from their spending and payment patterns, education, credit history, age, and marital status. The dataset contains $n = 30000$ individuals and we transform the labels so that $y_i = -1$ if a person will default on an upcoming credit card payment and $y_i = 1$ otherwise. In our case, we can have diverse separate contexts: 1) recourse cost considers all features and 2) recourse cost considers only actionable features, i.e. those related to spending and payment patterns and education, or 3) recourse cost considers only unjustified negative credit decisions (false negative).

As mentioned before, we can represent our beliefs as hyperparameters on the prior parameterized by the 1) noise scale $(\sigma)$ and 2) the regularization weights of the parameters $(\lambda)$. The noise scale measures the strength the belief that incoming credit decisions are randomly made while the regularization weights $\lambda$ represent a prior belief on how much features should be considered when making credit decisions. Generally, it is observed that as regularization increases, the number of individuals with recourse decreases
as regularization reduces the number of actionable features.

\subsection{Model Calibration}

To calibrate both $\lambda$ and $\sigma$, we formulate the search over different possible weights of $\lambda, \sigma \in [0.001, 0.01, 0.1, 1, 10]$ in an multi-objective optimization over with metrics that measure the consequences of such beliefs in the multiple contexts that we operate in. For the simplest setup, we consider $\lambda$ as shared across all features and since we consider two metrics: log likelihood, a measure of predictive accuracy, and recourse cost, a measure of the minimal amount of effort needed to receive a positive classification with decent probability. We emphasize that our metrics are now inherently probabilistic, which highlights the power of our model to incorporate belief uncertainty to generate an outcome and recourse distribution. 

Mathematically, our posterior distribution is given by $\mathcal{N}(w_{post}, A^{-1})$ which represents the diversity of beliefs arising from epistemic uncertainty. Then for an individual feature $x$, we note that $w^\top x \sim \mathcal{N}(w_{post}^\top x, x^\top A^{-1}x)$ is the predictive distribution. By using the sign as the decision threshold, the probability of receiving a positive outcome is given by the standard normal CDF of $\frac{w_{post}^\top x}{ \sqrt{ x^\top A^{-1}x}}$. Due to the low uncertainty estimates, we cap the log probability of wrong predictions at a minimum of $-5$ to prevent outliers from having exaggerated effects on aggregated statistics.

We fix our policy so that a positive credit decision is made if and only if $w_{post}^\top x \geq 0$, or equivalently if there is a positive outcome with at least $50\%$ chance over all choices of the model weights. Thus, while each parameter provides a different recourse, we can define recourse to be the minimum norm of a correction needed to receive a positive classification with respect to our fixed policy. In this case, this is equivalent to solving the following optimization $cost = \min \|c\| \text{  s.t.  } w_{post}^\top (x+c) \geq 0$. Note that this is $0$ when $w_{post}^\top x \geq 0$.

\begin{figure}[tbh]
\centering
\includegraphics[width=\columnwidth]{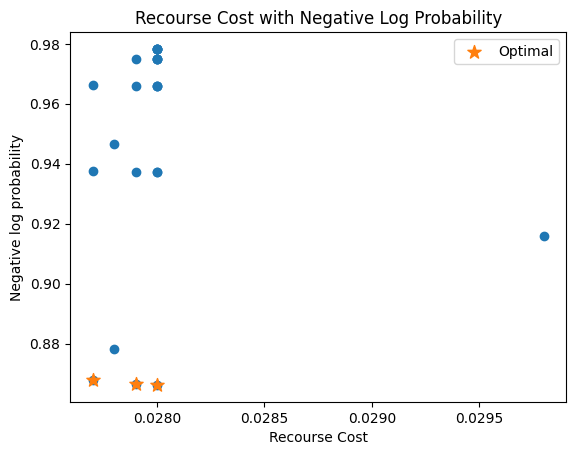}
\resizebox{0.8\columnwidth}{!}{%
\begin{tabular}{cccc}
\toprule 
$\sigma$ &  $\lambda$ & Recourse Cost &  Log-Prob \\
\midrule
10 & 0.1 & 0.0277 & 0.878 \\
10 & 1 & 0.0279 & 0.865 \\
10 & 10 & 0.028 & 0.866 \\
\bottomrule
\end{tabular}
}
\caption{Scatter plot of average recourse cost with negative log probability, along with a table of Pareto optimal points. Note that while many priors produce dominated points, there is a small Pareto frontier, which represents our set of optimal beliefs, and they have high confidence in this context.}
\label{fig:simple}
\vspace{-4mm}

\end{figure}

Our empirical results (see Figure~\ref{fig:simple}) show that there are many belief priors that produce suboptimal performance, as measured by our metrics. Concretely, we note that while we tested 25 discrete beliefs, there are only three that are Pareto optimal, specifically they are $\sigma = 10, \lambda = 0.1, 1, 10$. In fact, we emphasize that although all the Pareto optimal points share approximately the same negative log probability, the regularization value that produces the lowest recourse cost is $\lambda = 0.1$. From the Pareto frontier, it may seem that we agree with the conventional wisdom that as regularization increases, the average recourse cost increases since regularization reduces the number of actionable features. However, we note that the point with the highest recourse cost actually occurs when $\sigma = 10$ but $\lambda = 0.001$, which is perhaps surprising. These counter-intuitive conclusions highlight the impact and necessity of including counterfactual analysis in belief calibration.

\subsection{Contextual Calibration}

Appropriate definitions of recourse cost can vary significantly based on context. For example, we can use a feature-wise recourse cost to have lower costs for the actionable features, which are those related to spending and payment patterns and education. Specifically, let $D$ be a diagonal matrix such that $D_{ii} = 1$ for actionable features but $D_{ii} = 100$ is large for non-actionable features. Then, the cost of recourse is solved by $cost(D)= \min \|Dc\| \text{  s.t.  } w_{post}^\top (x+c) \geq 0$. We call this the actionable recourse cost. Accordingly, we also update our prior beliefs so that $\Sigma_p = D$, implying that we put higher emphasis on beliefs that non-actionable features should have near-zero weight and $\lambda$ tunes the strength of this prior.

\begin{figure}[tbh]
\centering
\includegraphics[width=\columnwidth]{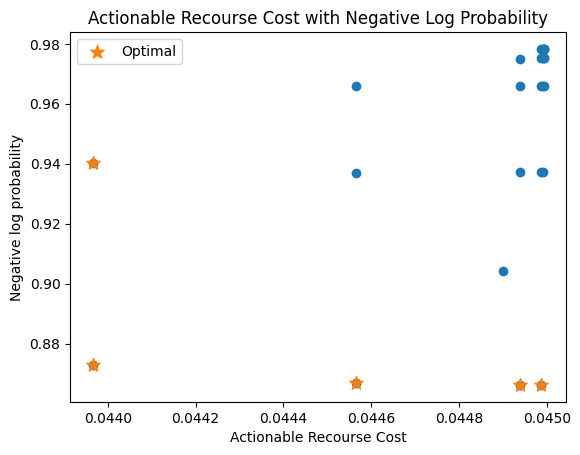} 

\begin{tabular}{cccc}
\toprule 
$\sigma$ &  $\lambda$ & Actionable Cost &  Log-Prob \\
\midrule
1 & 0.001 & 0.0439 & 0.94 \\
10 & 0.01 & 0.044 & 0.878 \\
10 & 0.1 & 0.0446 & 0.877 \\
10 & 1 & 0.0449 & 0.866 \\
10 & 10 & 0.045 & 0.866 \\
\bottomrule
\end{tabular}

\caption{Scatter plot of average actionable recourse cost with negative log probability, where non-actionable features have substantially higher cost. Most of the Pareto frontier has not changed, however the strength parameters that gives rise to the lowest actionable recourse cost has a decreased value of $\sigma,\lambda$ than before, justifying our argument that different contexts introduces more variance and uncertainty in Pareto optimal values of $\sigma,\lambda$.}
\label{fig:diag}
\vspace{-4mm}
\end{figure}

In this context, we see that from Figure~\ref{fig:diag} that while most of the Pareto points are relatively unchanged, the belief that corresponds to the lowest actionable recourse cost now becomes $\sigma =1, \lambda = 0.001$, as opposed to it being essentially $\sigma = 10$ from the previous analysis. In fact, in the previous context,  $\sigma = 10$ holds for all Pareto optimal points, implying that there is a strong (and justified) belief that many of the credit decisions are noisy. 
However, by adding an additional context, this increased the epistemic uncertainty over the optimal values of $\sigma$ as the much milder belief of $\sigma 
 = 1$ is also now a candidate.

Another context that is particularly relevant for sensible alignment is to distinguish two forms of recourse cost by utilizing a novel split of the cost of remediating negative credit decisions into those that are justified (true negative) and those that are unjustified (false negative). While decreasing actionable recourse has been extensively studied, especially when considering unjustified credit denials, we posit that it is perhaps equally important to uphold sufficiently costly consequences for justified credit denials in order to achieve societal alignment that disincentivizes risky behavior. Specifically, we define the false negative cost as equal to the recourse cost when the label is in fact $1$ and the cost is $0$ otherwise. Analogously, we define the true negative cost and our main objective is to minimize the false negative cost, while maintaining an adjusted minimum on the true negative cost, a threshold that we empirically set at $0.0188$. Our calibration results are shown in Figure~\ref{fig:tn} and we find that only two reasonable values of $\sigma, \lambda$ are proposed.

Overall, we observe that introducing new, yet related, contexts do not create substantially wild variation in beliefs, but simply proposes a mild re-calibration, or a strict subset of common belief settings that generalize across contexts.. Specifically, note that some of our previously optimal beliefs are still on the altered Pareto frontier under diverse contexts (note also that when adding a new context as an additional objective, it will not remove points on the Pareto front). This implies that our beliefs on relative prior strengths generalize across different contexts, but introducing differing costs can justify adding another belief prior into our set of non-dominated beliefs. However, we note that belief calibration can differ significantly given new counterfactual information.

\subsection{Policy Calibration}

Model design and calibration is only one of multiple processes that can be aligned in the credit outcome decision. In this section, by introducing subjectivity, we add a new hyperparameter $\beta$, that captures beliefs in policy design, which takes a model output to outcomes and recourses. Just as in Bayesian optimization, we can exploit our model's uncertainty predictions to encode the belief of giving individuals the ``benefit of the doubt" by giving a positive outcome when $\mu(x) + \beta \sigma(x) \geq 0$, where $\mu, \sigma$ are the predicted mean and standard deviation. Note that previously our policy is setting $\beta = 0$ and therefore, our more lenient policy is equivalent to allowing easier credit when there is higher uncertainty or doubt in the model.

Mathematically, we now define recourse to be a slightly complicated quadratic optimization of the form $cost(\beta) = \min \|c\| \text{  s.t.  } w_{post}^\top (x+c) + \beta(x+c)^\top A^{-1}(x+c) \geq 0$. This requires an eigenvector computation and is significantly more computationally expensive; therefore we use a subset of the dataset, around $1000$ individuals, to perform our experiments. Furthermore, we note that the log-likehood calculations now include $\beta$ term, to reflect the greater leniency to generating a positive outcome. We let $\beta \in [0.1, 1, 10]$, where a larger $\beta$ represents easier credit.

\begin{figure}[tbh]
\centering
\includegraphics[width=\columnwidth]{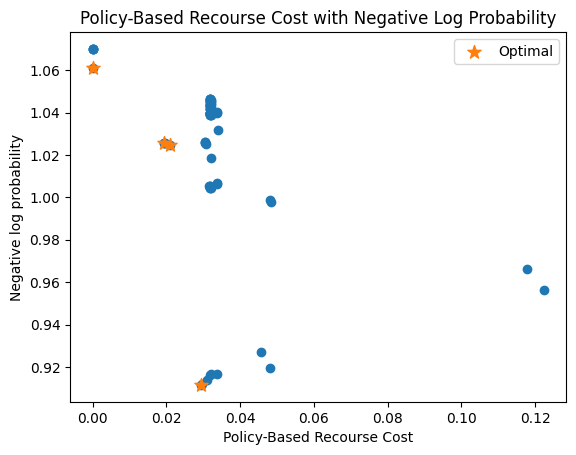}
\resizebox{0.9\columnwidth}{!}{%
\begin{tabular}{ccccc}
\toprule 
$\sigma$ &  $\lambda$ & $\beta$ & Policy Cost &  Log-Prob \\
\midrule
10 & 0.001 & 10 & 0.0 & 1.06 \\
1 & 10 & 10 & 0.019 & 1.025 \\
1 & 0.01 & 10 & 0.021 & 1.024 \\
10 & 10 & 1.0 & 0.029 & 0.911 \\
\bottomrule
\end{tabular}
}
\caption{Scatter plot of average policy-based recourse cost with negative log probability, along with a table of Pareto optimal points. Note that all the Pareto frontiers do not use the small values of $\sigma, \beta$, meaning that calibrating our beliefs to recognize that credit decisions are inherently noisy and adding some leniency into the final decision policy produces more accurate probabilistic predictions while also promoting better societal outcomes.}
\label{fig:policy}
\vspace{-4mm}

\end{figure}

From our results, we find the surprising fact that giving more lenient credit decisions can actually help both accuracy while reducing policy-based recourse cost (see Fig~\ref{fig:policy}). Specifically, the Pareto optimal beliefs are calibrated to have high values of $\sigma, \beta$, implying that higher uncertainty in credit modeling and using those uncertainty measures in giving more lenient credit decisions improve our predictive calibration, as the most accurate model has a $\beta = 1$ and $\sigma = 10$.     

\section{Conclusion}


As we continue to build decision systems modeled after human beliefs in high stakes contexts, it is critical to acknowledge the diversity of human beliefs, rising from both population-level subjectivity but also individual-level epistemic uncertainty, and the difficulty that such diversity presents, which we term the meta-alignment problem. To tackle the challenge of calibrating beliefs to different contexts, we proposed to leverage counterfactual reasoning to not only consider measures of predictive success but also social benefit in the form of outcomes and recourses in a ``belief calibration cycle.'' We presented experiments on a credit decision dataset, leveraging Bayesian modeling to analyze how uncertainty in beliefs on the hyperparameter values leads to a distribution of outcomes and recourses, which in turn informed the optimal choices in beliefs for obtain the best outcomes and recourses. By doing so, we find that belief calibration leads to surprising conclusions, such as the case for increased leniency in credit applications, which highlights the importance of utilizing counterfactual analysis in belief calibration.

\bibliographystyle{abbrvnat}
\bibliography{ref}



\end{document}